\documentclass{article} 
\usepackage{CoLLAs2022,times}
\usepackage{graphicx}
\usepackage[margin=3cm]{geometry}
\usepackage{algorithm2e}
\SetKwInOut{Parameter}{parameter}

\RestyleAlgo{ruled}
\SetKwComment{Comment}{/* }{ */} 
\SetKwInput{Input}{Input}
\SetKwInput{Initialize}{Initialize}
\SetKwInput{Output}{Output}
\SetKwFunction{Average}{Average}
\def\algo{
    \begin{algorithm}[hbt!]
    \caption{Training protocol for ${i}^{th}$ incremental task.}\label{alg:two}

    \Input{\\
        \Indp
        $D_{i}$: Dataset for current task $i$ \;
        $W_{i-1}$ (if $i > 1$): Trained weights of previous task\;
        $V_{i-1}$ (if $i > 1$): Eigenvectors of previous task\;
    }
    
    \Initialize{\\
        \Indp
        \eIf{$i = 0$}{
            Randomly initialize $W_{i}$ \;
        }{
            \For{l in layers}{
                Randomly initialize $W^{l}_{trainable}$ for each layer \; 
                Freeze weights of the previous task $W^l_{i-1}$ \;
                Initialize $W^{l}_{i} \leftarrow W^l_{i-1} + V^l_{i-1} * W^{l}_{trainable}$ \;
            }
        }
    }
    
    \For{epoch}{
        \For{mini-batch}{
            Forward pass through $W_{i}$ for $D_{i}$\;
            Compute task specific loss $L_{BCE_{i}}$ for $D_{i}$ \; 
    }
    }
    \For{each-train-sample}{
        \For{each-layer}{
            $C^l = \frac{dL_{BCE}}{dh^l}^t \frac{dL_{BCE}}{dh^l}$ \;
        }
    }
    \For{each-layer}{
        Compute Sorted Eigenvector $V^l_{i}$ on \Average($C^l$) \;
    }
    
    \Output{\\
        \Indp
        $W_{i}$: Trained weights of previous task\;
        $V_{i}$: Eigenvectors of previous task\;
    }
    
    \end{algorithm}

}

\usepackage{hyperref}
\hypersetup{
    colorlinks=true,
    linkcolor=red,
    filecolor=magenta,      
    urlcolor=magenta,
    citecolor=purple,
    pdftitle={Overleaf Example},
    pdfpagemode=FullScreen,
    }

\newcommand\blfootnote[1]{%
  \begingroup
  \renewcommand\thefootnote{}\footnote{#1}%
  \addtocounter{footnote}{-1}%
  \endgroup
}

\title{Gradient Correlation Subspace Learning against Catastrophic Forgetting}

\author{Tammuz Dubnov\textsuperscript{*} \\
Reichman University, Israel\\
\texttt{tdubnov@gmail.com} \\
\And 
Vishal Thengane\textsuperscript{*}  \\
Independent Researcher, India \\
\texttt{vgthengane@gmail.com} \\
}

\collasfinalcopy 

\begin{document}

\maketitle

\blfootnote{\textsuperscript{*} denotes equally contributed.}

\begin{abstract}
Efficient continual learning techniques have been a topic of significant research over the last few years. A fundamental problem with such learning is severe degradation of performance on previously learned tasks, known also as catastrophic forgetting. This paper introduces a novel method to reduce catastrophic forgetting in the context of incremental class learning called Gradient Correlation Subspace Learning (GCSL). The method detects a subspace of the weights that is least affected by previous tasks and projects the weights to train for the new task into said subspace. The method can be applied to one or more layers of a given network architectures and the size of the subspace used can be altered from layer to layer and task to task. Code will be available at \href{https://github.com/vgthengane/GCSL}{https://github.com/vgthengane/GCSL}


\end{abstract}

\section{Introduction}
Traditionally, as a neural network learns multi-class classification, the network learns to extract features indicative of the target labels and to perform the classification. 
When learning on all target labels simultaneously, networks are able to reach the highest accuracy presumably because they can learn the features relevant to all labels at the same time. If a network is trained on a set of labels and then at a later point trained on a different set of labels, the network often "catastrophically forgets" how to classify the original labels. This field is known as Continual Learning and Lifelong Learning. In the process of learning new labels, it may need to learn new features that may overwrite the previous fragile weights that were able to extract the appropriate features and perform classifications for the previously-learnt labels. 

Artificial neural networks cannot continuously adapt to dynamic environments with changing demands but are limited to high performances only in the original environment and labels they were trained on. This field of research, Continual Learning, is growing as increasingly complex deep learning networks are trained on sizable databases using significant computational resources. To many it would be very attractive to be able to utilize previously trained deep networks when an additional task requirement arises, with the previous tasks still required. Currently, it is often needed to retrain a network from scratch to reach the highest performance or perform fine-tuning that may decrease the performance on previously learnt tasks if they are not fairly represented in the training set. 

\begin{figure}[ht]
\begin{center}
\includegraphics[width=\linewidth]{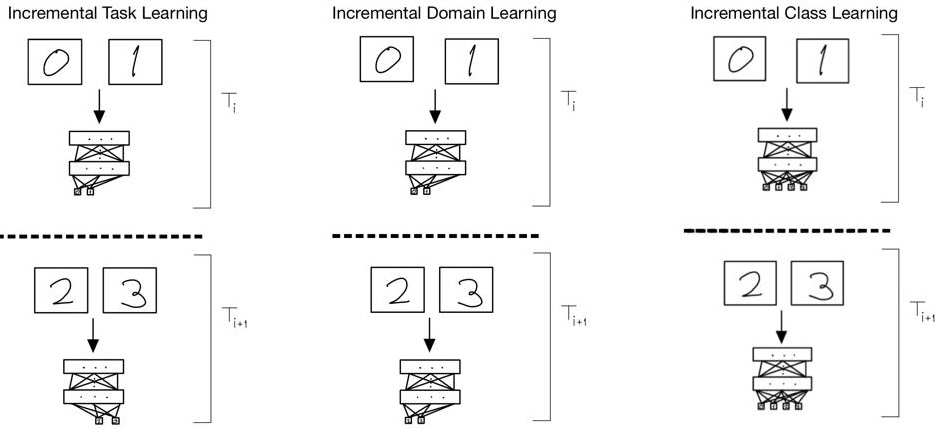}
\caption{The three continual learning scenarios for the first two tasks of the MNIST dataset. The top shows the first task and the bottom shows the task following it. The pattern in the final logit layer for each learning type is extended for $T_{j>(i+1)}$ tasks. \citep{hsu2019reevaluating}}
\label{icml-historical}
\end{center}
\end{figure}

There are three scenarios of incremental learning, as can be seen in Figure 1.
Incremental Task Learning attaches a new set of neurons for new tasks it learns allowing a set of neurons to uniquely learn the probability distribution of the labels at hand without affecting neurons in the final layer responsible for classifying previously learned labels. Incremental Domain Learning uses the same logits for multiple tasks such that $L_{1}^{1}$ (label 1, the subscript, from task 1, superscript) and $L_{1}^{2}$ (label 1 from task 2) would activate the same logit limiting the overall network to only be able to classify between $L_{1}^{t}$ and $L_{2}^{t}$ and not $L_{x}^{1}$ and $L_{x}^{2}$ for some $x$. For example, if learning over the MNIST dataset where each task trains on the pair of classes in Figure 1, domain learning would enable differentiation broadly between even and odd numbers but not between the various even numbers. This forces the network to map different probability distributions to the same labels which may not be ideal for all use-cases. Incremental Class Learning works with a predefined maximum amount of labels it can potentially classify, and learns each task and label probability distribution to the next available logit. This paper focuses entirely on Incremental Class Learning which is also known to be the most challenging scenario \citep{hsu2019reevaluating}.

This paper introduce a method to deal with incrementally growing set of classes by continuously calculating the gradients correlation matrix and learning on weights which are spanned by least significant eigenvectors. The method ideally leads to learning new tasks with as little interference on previously learned tasks as possible. The method propagates from one task to the next for each layer on the trained weights, and calculates the set of eigenvectors of the gradients correlation matrix of the loss with respect to the activation. When training a new task, the previous weights are frozen, and new trainable weights of the same structure are initialized. 
When training a new task, the training happens on new trainable weights projected on the subspace, defined by a selected number of eigenvectors corresponding to the smallest eigenvalues, through the sum of the trainable weights with the frozen weights from before the current training. This paper demonstrates this method using the MNIST dataset of handwritten digits \citep{726791}, and the Fashion MNIST dataset of article images \citep{xiao2017_online}.

\section{Related Works}
Within continual learning, the wide range of relevant use-cases has led to the development of different types of methods to address the problem of catastrophic forgetting. Approaches from architecture expansion, to regularization and replay methods, are suited for different scenarios with different limitations. If computational resources limit the size of the network then a architecture expansion-based approach may not be appropriate. If memory or data privacy based limitations prevent samples from task datasets from being stored after training then the replay methods may not be appropriate. Broadly speaking, there are the following approaches:

\subsection{Architectural methods}
Architectural methods rely on expanding the size of the network to focus the training of new tasks on new or less-important sub-networks of the network and therefore minimize the forgetting of previous tasks on critical portions of the network. The Progressive Neural Network \citep{DBLP:journals/corr/RusuRDSKKPH16} method freezes the base model, able to perform previous tasks, and adds sub-networks for each new task. Methods such as Dynamically Expandable Networks \citep{DBLP:journals/corr/abs-1708-01547} expand the network by splitting and duplicating important sub-networks for new tasks to learn, and the Reinforced Continual Learning method \citep{DBLP:journals/corr/abs-1805-12369} searches for the best neural architecture for the new task using reinforcement learning. There are approaches that work to reduce the size of the growing network after each learning. Such methods may add compression step after every learning of a new task, such as Adding Multiple tasks to a Single Network by Iterative Pruning \citep{https://doi.org/10.48550/arxiv.1711.05769}, or use distillation to return to the original network size, such as Progress and Compress: A scalable framework for continual learning \citep{https://doi.org/10.48550/arxiv.1805.06370}. While these methods may be effective, it is a challenge to mitigate the growing network size after each new task learned. 

\subsection{Replay methods}
Replay methods rely on training using replays of previous tasks while training on a new task. Replays of previous tasks can be samples from past tasks that are stored for future training, or generated samples that are based on learned networks that can generate examples of the same distribution as the previous tasks' training set. Sophisticated methods such as Meta-Experience Replay \citep{DBLP:journals/corr/abs-1810-11910} have been proposed that limit the amount of samples that need to be stored over time by incorporating optimization based meta-learning. Generative methods have been proposed such as Deep Generative Replay \citep{DBLP:journals/corr/ShinLKK17} that also train GANs to generate examples like the current task's training set to interleave into the training of future tasks. These methods rely on the availability of additional memory, either for the samples from previous sets or for storing the GAN networks, so they may not be appropriate in every scenario.  

\subsection{Regularization-based methods}
Regularization-based methods rely on sophisticated methods to force the training to occur in such a way that it forgets the previous tasks as least as possible. The Elastic Weight Consolidation \citep{DBLP:journals/corr/KirkpatrickPRVD16} calculates the Fisher information matrix after each task training and limits the amount of learning for future tasks that can be done on parameters that are considered important to previous tasks due to their high information score. 
The method proposed in this paper is within a subset of the regularization-based methods that incorporate gradient calculations into their approach. Numerous methods exists within this domain such as Orthogonal Gradient Decent (OGD) \citep{DBLP:journals/corr/abs-1910-07104}, Orthogonal Weight modulation (OWM) \citep{DBLP:journals/corr/abs-1810-01256}, Gradient Episodic Memory \citep{https://doi.org/10.48550/arxiv.1706.08840} and Gradient Projection Memory (GPM) \citep{DBLP:journals/corr/abs-2103-09762}. For each task learned, OGD stores the gradient directions in memory so that new tasks can learn in the orthogonal directions to the gradients of previously learned tasks. Whereas OWM modifies the weights of the network with projected metrics that it stored from past tasks, calculated using recursive least squares. GEM uses episodic memory to project gradients to utilize shared features across tasks. And lastly, GPM uses episode storage to directly alter the gradients during back-propagation by subtracting the gradient times the space of significant representation, from previous tasks, from the gradient when updating the weights.

\subsection{Differences from similar approaches}
OWM, GEM and GPM are closely related to the method proposed in this paper. Unlike the approaches mentioned, the GCSL method proposed here is based on working in the orthogonal subspace to the calculation of the correlation matrix of the gradients for the newly learned task. GCSL works in the weight space rather than the gradient space. The proposed method requires that a subset of the eigenvectors of the correlated matrix be stored alongside the actual network. The method temporarily initiates new weights to use with said eigenvectors and freezes the original weights and back-propagates regularly; unlike GEM and GPM that affect the gradient calculation step directly during learning. Unlike OWM, the proposed method uses correlation matrix to try to capture the relationship between the logits in the layer with respect to the loss in order to work in a subspace that least affects the relationships between all the logits in the layer. The GCSL  method of freezing the weights from the previous tasks and adding new weights for the current learning, working in the weight space rather than the gradient space, allows the method to be run with any optimizer.



\section{Gradient Correlation Subspace Learning}

\algo

Like other gradient-based approaches, GCSL involves extra steps after training on a new task in order to capture a subspace that will be orthogonal to the task currently learned. The method proposed works with any fully connected layer and the strength per layer where it is applied can be set for every new task. An overview of the approach can be seen in Algorithm 1. The method does not limit or affect the optimizers that can be used while training.

\subsection{Subspace Learning}

When propagating an input forward through the layers of the network, broadly speaking, each layer extracts features and/or combines features from the output of the previous layer to perform the overall task of the network. After training, specific features are learned in each layer. When learning a new task with the intention of retaining the previous tasks, it is desirable to mitigate the effect of the learning on the ability to perform previous tasks learned. A strategy to do so is by finding a subspace that least affects the gradient of previously learnt tasks for each layer and only performing learning within that subspace.

Given a hidden layer $l$ of size $n_l$, the activation at task $t_{k-1}$ are given by $$h_{t_{k-1}}^l = g(W^l_{t_{k-1}} x^l)$$ with $g()$ being the activation function (defaulted to RELU), $W^l_{t_{k-1}}$ the weights matrix of size $[n_{l-1} \times n_l]$, and $x^l$ the inputs from a previous layer of size $n_{l-1}$. In order to find a subspace of weights that is the least effected during learning of the previous task, the correlation matrix of the learning gradients is computed as $$c_g^l = \frac{dL}{dh^l}^t \frac{dL}{dh^l}$$ for a given loss function $L$ for a specific sample of the training set from the current task. During each new task learning, the training set includes numerous examples so an accumulation of the average value of $c_g^l$ is stored into a matrix $C_g^l$.
Now we define $V = \{v_1, ..., v_{n_l}\}$ to be the set of eigenvectors of $C_g^l$ with eigenvalues $\{\lambda_1, ..., \lambda_{n_l}\}$, ranked from smallest to largest. The eigenvectors are computed using PCA and are therefore orthonormal. Taking the $n$ smallest eigenvalues with $n<n_l$, we define a projection operator over the weights $W^l_{t_{k-1}}$ as the matrix $V=[v_1,v_2,...v_n]$ of size $[n \times n_l]$. 

In order to constrain the learning to be within the subspace, we define a new set of weights $W^l_{trainable}$. In the implementation of the experiments below, a Xavier initialization \citep{pmlr-v9-glorot10a} was applied. The weights of the previous tasks $W^l_{t_{k-1}}$ are frozen and $W^l_{trainable}$ is updated during the $t_k$ training session. 
During training session $t_k$, the $V^l_{t_{k-1}} W^l_{trainable}$ is attached parallel to the weight $W^l_{t_{k-1}}$. At the end of the training session, the two weight matrices can be simply combined as follows:  $$W^l_{t_k} = W^l_{t_{k-1}} + V^l_{t_{k-1}} W^l_{trainable}$$ such that inference for tasks $t_1$ to $t_k$ is  easily run using  $$h_{t_k}^l = g(W^l_{t_{k}} x^l)$$, which we can use for inference of tasks $t_1$ to $t_k$. 
Before continuing to training $t_{k+1}$, $C_g^l$ is calculated for the training set or a subset of the training set. $V^l_{t_{k}}$ is then calculated using $C_g^l$, with $V^l_{t_{k}}$ being the $n$ eigenvectors with the largest corresponding eigenvalues where the desired $n$ can be defined and changed between learning tasks. We discuss the heuristics for determining $n$ for each layer in the experimental section.

\subsection{Task-Specific Binary Cross Entropy Loss}
As the network learns tasks $t_k$ for $k>1$, the network only has access to samples with the labels from task $t_k$ but should ideally keep the ability to properly label inputs from previous tasks $t_{j<k}$. The traditional loss function for multi class learning is Cross Entropy loss (CE Loss). The limitation with the default CE Loss, that factors in the loss of previously learned labels, would effectively teach the network to mute logits for those edges after enough training iterations where the labels of those tasks have not been seen. Instead, a Binary Cross Entropy loss (BCE Loss) is used, the BCE loss is defined only for the current training session. For task $t_k$, the BCE Loss for each label is defined as $$loss(i,y) = y \log(i) + (1-y)\log (1-i)$$ where $y$ is the target labels of task $t_k$. Note that the definition BCE loss does not limit the number of labels in $y$ from being of any length. The BCE loss effectively prevents the propagation of the loss through logits of labels from previous tasks that are not represented in the training set of the current task $t_k$. 

\subsection{GCSL Steps}
As can be seen in Algorithm 1, GCSL consists of three main steps. The first step consists of training on task $t_i$, where if $i=0$ then the network is trained normally and if $i>0$ then new weights $W^l_{trainable}$ are defined for layers for which GCSL is applied such that the training occurs with $W^l_i$ defined as $W^l_{i-1} + V^l_{i-1} W^l_{trainable}$ with $W_{l_{i-1}}$ frozen and only $W^l_{trainable}$ updated. The second steps involves calculating the accumulative correlation matrix for each layer over all of task $t_i$'s training set or a subset of the training set. The last step calculates the eigenvectors $V^l_i$ for every layer, sorted by the corresponding eigenvalues. From task to task, $W_i$ and a subset of $V_i$, where the subset size may vary per layer, are the only elements that need to be used for the next task training.

\section{Experimentation}
To properly explore the GCSL method, we run two types of experiments. The GCSL Experiments focuses internally on the different configurations possible within the method and compares the results for different parameters configurations in GCSL. The second set of experiments, the Comparison Experiments, uses a static configuration and compares the results to a state-of-the-art method GPM to gauge the promise of the method. 

\subsection{Datasets}
The experiments where performed on two different datasets, the MNIST \citep{726791} and Fashion MNIST datasets \citep{xiao2017_online}. MNIST is a dataset of handwritten digits where as Fashion MNIST is a dataset of article images. Both datasets have the same training size of 60k, with the same train/test split as well, and are both made up of images of 28x28 pixel.

\subsection{Network Architecture Considerations}
\subsubsection{GCSL Experiments}
The experiments were designed to challenge the tasks as much as possible and validate that orthogonal gradient spaces are indeed being found rather than unused areas of the network being used for the tasks. As such, each dataset was empirically tested to find the smallest model size at which the baseline results could be achieved at close to the highest mark possible. The baseline task was to train a network to learn to classify all tasks at once, rather than incrementally. This baseline serves as the true upper bound to the top performance a network can perform on the datasets. The final model is a fully connected network that receives a flattened 28x28 image followed by hidden layers that lead to an output of 10 nodes. The MNIST dataset was run with two hidden layers of size 20 each and the Fashion MNIST dataset was run with a hidden layer of size 40 followed by a hidden layer of size 20. The experiment training was done on the label pairs in the following order (0,1), (2,3), (4,5), (6,7), (8,9). The batch size during training is 128 and each task was learned over the same amount of epochs. For MNIST each task was trained for 10 epochs and for the Fashion MNIST each task was trained for 5 epochs. 

\subsubsection{Comparison Experiments}

The Comparison Experiments were designed to compare the performance of GCSL with GPM in different scenarios with different architecture sizes. As such, multiple iterations of each architecture was trained. The experiment results show the results over all the iterations. Each iteration has a different seeds and the order and pairing of the labels learned at each task are also randomized before the next iteration. The batch size during training is 128 and each task was learned over 5 epochs.




\subsection{Validation Set}
As the experiment proceeds to tasks $t_2$ to $t_5$, an accumulative validation of all tasks $t_{i\le k}$ is aggregated to evaluate the overall  The final accuracy represented in the results below shows the accuracy over the validation set of all tasks combined.

\subsection{$V$ Layer Configuration}
\subsubsection{GCSL Experiments}
For each dataset seven experiments were run; each experiment consisted of 25 runs. A baseline experiment that trained the task regularly, not incrementally but trained all labels together in a single task. A bland incremental learning experiment that did not use the GCSL method but simply trained one task at a time and faced catastrophic forgetting; marked as ${L_1:0, L2:0}$. Three more experiments were run, one using the largest subspace size with a layer configuration for the entire $V$, a second that used half of the possible configuration size for each layer, and a third that used a quarter of the possible configuration size. Two more experiments were conducted that used the configuration from the top of the last three experiments. The last two experiments isolate the layer configuration to just one layer at a time. 

For the MNIST dataset, the network architecture is [20,20] (two hidden layers of size 20) and the full configuration used $V$ layer configuration of [20,20], the second used a configuration of [10,10], and the third used a configuration of [5,5]. The two last experiments used a configuration of [10,] and [,10]. For the Fashion MNIST dataset, the network architecture is [40,20] and the full configuration used $V$ layer configuration of [40,20], the second used a configuration of [20,10], and the third used a configuration of [10,5]. The two last experiments used a configuration of [20,] and [,10].

\subsubsection{Comparison Experiments}
To remain consistent when comparing different architecture sizes for multiple methods, the size ratio of $V$ was set constant per dataset. The experiment on MNIST utilized a ratio of 80\% whereas the experiment on FMNIST utilized a ratio of 50\%.

\subsection{Training Parameters}
\subsubsection{GCSL Experiments}
The MNIST experiments were run with an SGD optimizer at a learning rate of 0.05 for 15 epochs at each task. The Fashion MNIST experiments were run with an Adam optimizer at a learning rate of 0.002 and betas of 0.9 and 0.999. Each experiment was conducted over 25 runs with a different random seed for each run. The results show a box plot of the accuracy over the aggregated validation sets over the 25 runs for each configuration.

\subsubsection{Comparison Experiments}
Experiments were run comparing the results of GCSL to GPM, a state-of-the-art approach with a comparable approach. The experiments were run with an SGD optimizer on both the MNIST and FMNIST datasets. The MNIST experiments trained with a learning rate of 0.01 and the FMNIST are trained with a learning rate of 0.02. Each experiment was conducted over 10 iterations of different runs with different label orderings and different seeds.

\section{Results}

\begin{figure}[ht]
\vskip 0.2in
\begin{center}
\centerline{\includegraphics[width=0.75\linewidth]{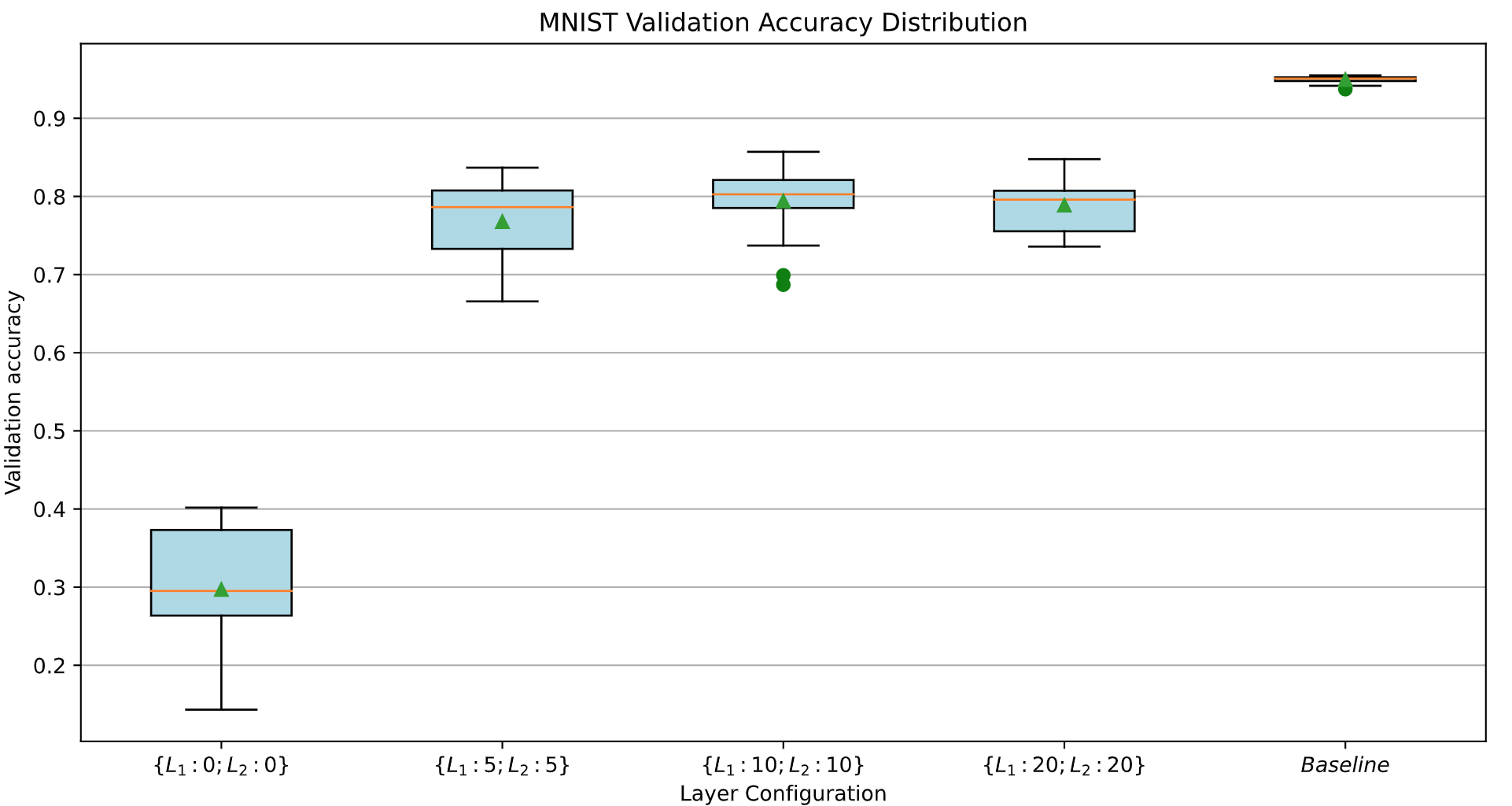}}
\caption{Results for different configuration sizes on all layers the MNIST dataset. The accuracy for the accumulative validation set at the end of the final task training. The $Baseline$ serves as an upper bound to the possible performance of the given network size, and ${L1:0, L2:0}$ serves as a lower bound for the worst possible performance by learning incrementally without the GCSL technique. Note that the ${L1:20, L2:20}$ is the largest possible gradient subspace configuration. }
\label{icml-historical}
\end{center}
\vskip -0.2in
\end{figure}

\begin{figure}[ht]
\vskip 0.2in
\begin{center}
\centerline{\includegraphics[width=0.75\linewidth]{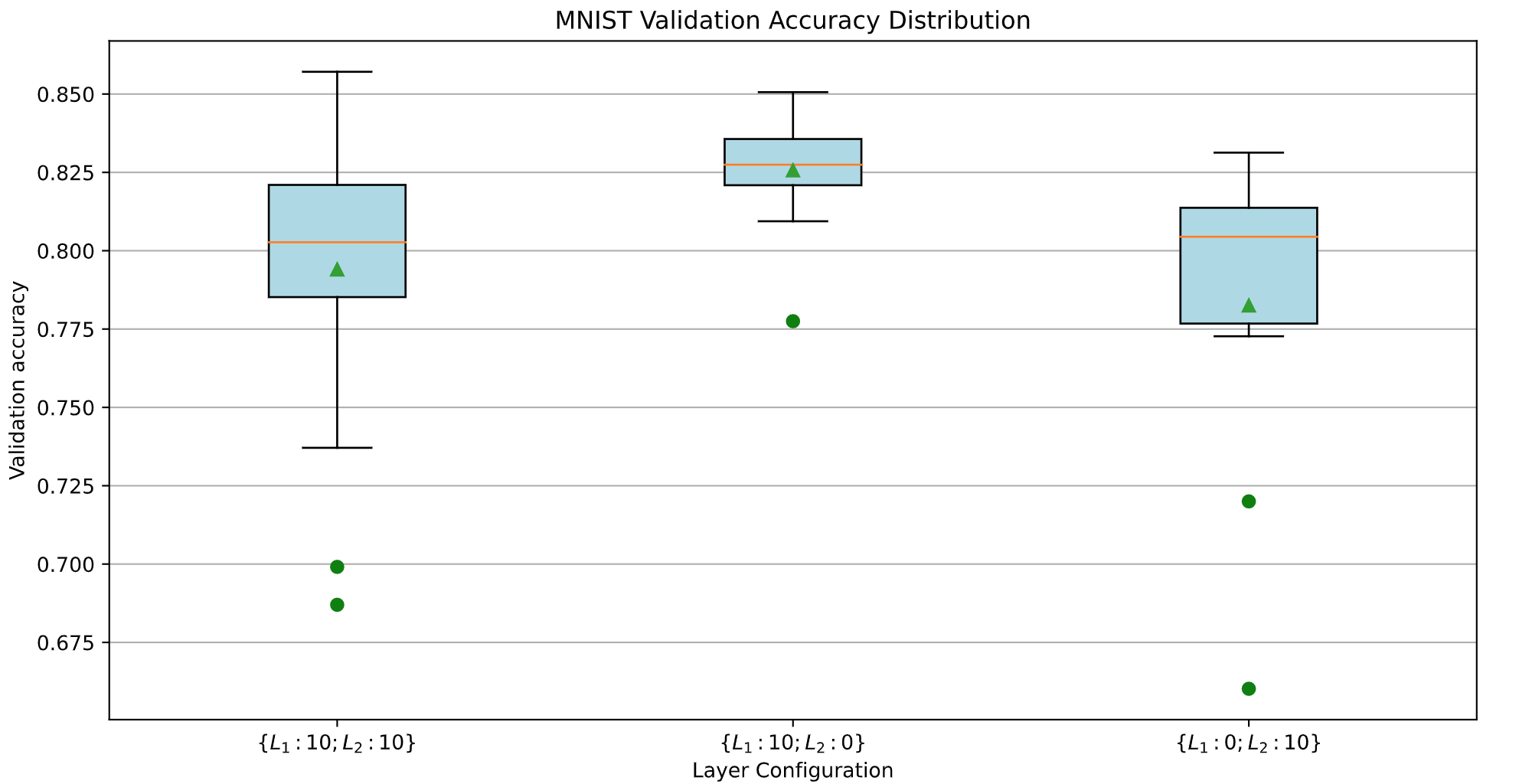}}
\caption{Results for different layers configuration using the size with the best performance from Figure 2 on the MNIST dataset. Note that the ${L1:10, L2:10}$ means the GCSL method was performed on both layers, ${L1:10, L2:0}$ means the method was performed on the first layer, and ${L1:0, L2:10}$ means the method was performed on the second layer. }
\label{icml-historical}
\end{center}
\vskip -0.2in
\end{figure}

\begin{figure}[ht]
\vskip 0.2in
\begin{center}
\centerline{\includegraphics[width=0.75\linewidth]{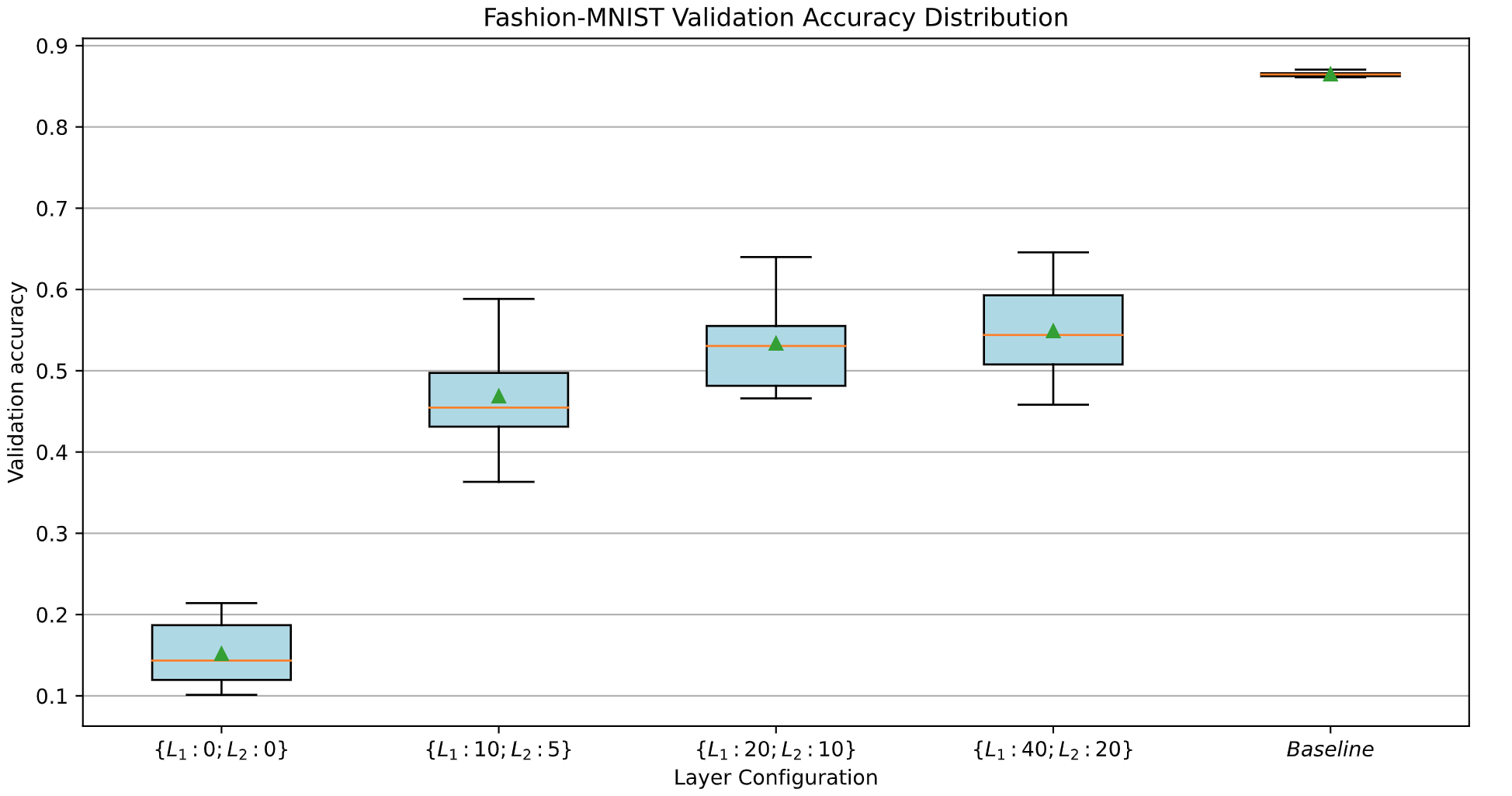}}
\caption{Results for different configuration sizes on all layers on the Fashion MNIST dataset. The accuracy for the accumulative validation set at the end of the final task training. The $Baseline$ serves as an upper bound to the possible performance of the given network size, and ${L1:0, L2:0}$ serves as a lower bound for the worst possible performance by learning incrementally without the GCSL technique. Note that the ${L1:40, L2:20}$ is the largest possible gradient subspace configuration. }
\label{icml-historical}
\end{center}
\vskip -0.2in
\end{figure}

\begin{figure}[ht]
\vskip 0.2in
\begin{center}
\centerline{\includegraphics[width=0.75\linewidth]{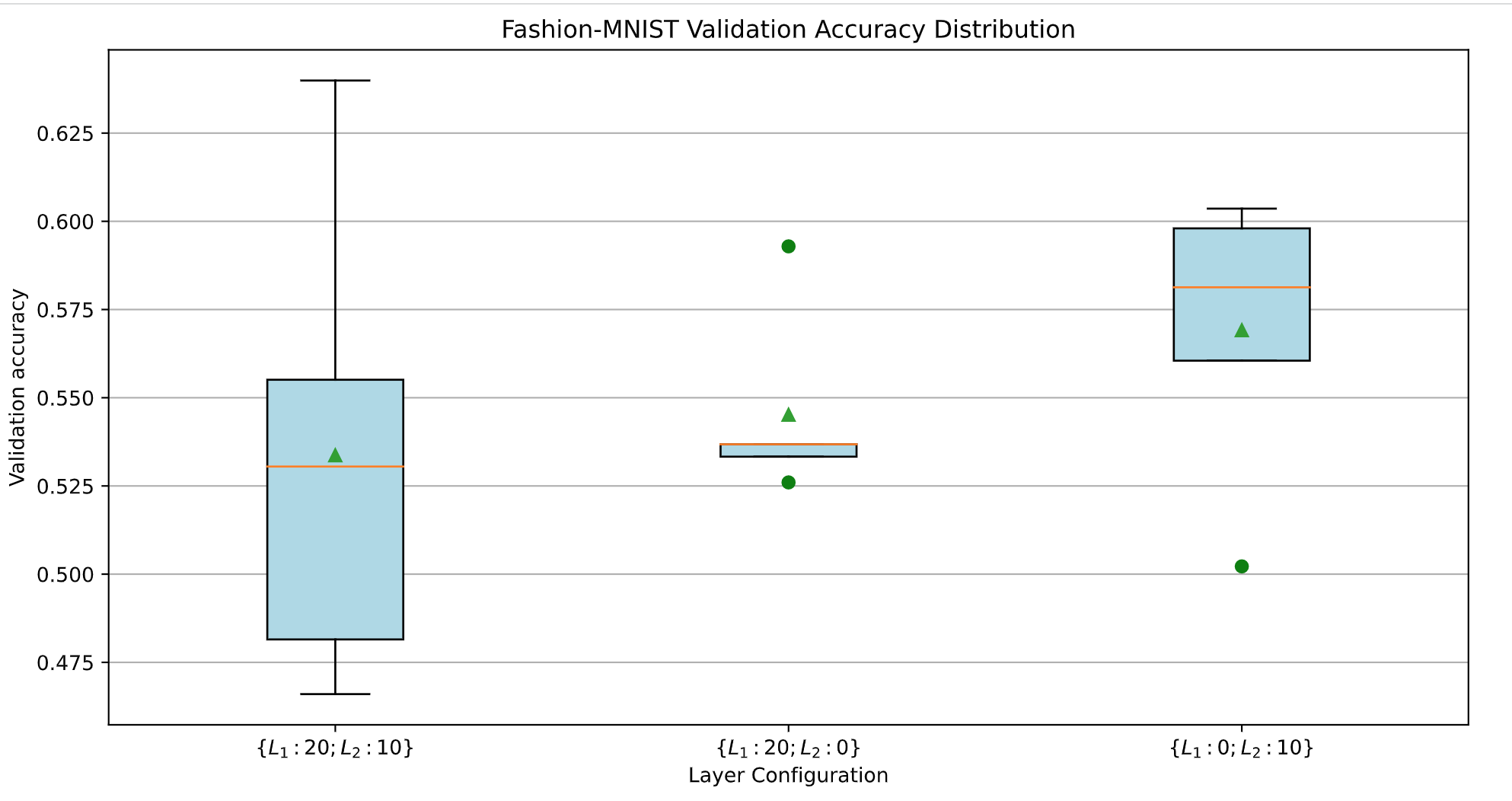}}
\caption{Results for different layers configuration using the size with the best performance from Figure 4 on the Fashion MNIST dataset. Note that the ${L1:20, L2:10}$ means the GCSL method was performed on both layers, ${L1:20, L2:0}$ means the method was performed on the first layer, and ${L1:0, L2:10}$ means the method was performed on the second layer. }
\label{icml-historical}
\end{center}
\vskip -0.2in
\end{figure}

\subsection{GCSL Experiments}

\subsubsection{Results for MNIST}
As can be seen in Figure 2, using the GCSL method improves the default catastrophic forgetting behavior by over 2.5 times (from an average accuracy of 30\% to a top average accuracy of 80\%). It can be seen that the largest possible subspace size, [20,20], does not immediately return the best results. Generally, the larger the subspace the more trainable weights are available during training but the more the subspace encroaches on the gradients of the previous tasks.

As can be seen in Figure 3, using the GCSL method on both layers does not necessarily lead to the best results. The top result for this dataset was using the method on the first layer, $L1$. When running layer 2, it is clearly visible that the results are significantly worse. 

\subsubsection{Results for Fashion MNIST}

As can be seen in Figure 2, using the GCSL method improves the default catastrophic forgetting behavior by over 3 times (from an average accuracy of 15\% to a top average accuracy of 55\%). The baseline wasn't able to reach a 90\% accuracy showing that this is a challenging task for the network. In this challenging task, the larger the subspace, the better the average accuracy. This generally means that for this dataset there is more pressure to learn new tasks than fear of forgetting previous tasks.

As can be seen in Figure 5, the best results are from the method only using layer 2, $L2$. Broadly speaking, the later layers are more involved in the task of the actual classification as compared to the earlier layers that are more involved in extracting features. It seems that for this network architecture and dataset, being able to learn over the earlier layer of feature extract but learn orthogonally on the later layer and minimize the deteriorating impact of learning new tasks on the classification segment, leads to better results. This would generally make people wonder if the features learned for the different labels may be similar and indicative of other labels and therefore be beneficial or at least not detrimental to continue to learn.

\subsubsection{Comparing the Datasets}
When comparing the results of the two datasets, the behavior of GCSL can be better understood. It appears that the Fashion MNIST is a harder task than the MNIST since the baseline results for the Fashion MNIST were significantly lower. For anyone familiar with the two datasets, even for a human the Fashion MNIST may be challenging. From the results of different layer subspace configuration sizes, it can be seen that the easier MNIST task can be learned with a subspace less than the maximum and in facts benefits from the decreased forgetting. Whereas the more nuanced and challenging task of the Fashion-MNIST needed a bigger emphasis on a larger sized subspace to have more room to learn the new tasks.

When comparing the optimal layer for GCSL per dataset, whether both, the first or the second layer, again the characteristics of the datasets and tasks play a big role. It may not be possible to note the reason, but either due to the relative simplicity of the MNIST dataset or due to the relatively distinct features required to recognize handwritten digits, the MNIST performs better when the method is applied to the first layer. This means that the first layer, that can be broadly thought to be more focused on feature extraction, benefits from the orthogonal subspace that such that feature learned from previous tasks are forgotten/overwritten as little as possible. Whereas directly updating the second layer, broadly thought to be more focused on the classification portion, performs better than applying the GCSL method on the layer. When looking at the Fashion MNIST dataset, the opposite is true. The network performs significantly better when the method is applied only to the second layer. This seems to indicate that obstructing and slightly altering the feature level is not damaging whereas the classification is much more sensitive. 

\subsection{Comparison Experiments}
The performance of GCSL performs on-par or above the performance of GPM as can be seen in tables \ref{MNIST-comptable} and \ref{FMNIST-comptable}. As the architecture size grows the accuracy of both methods improves, but on the MNIST dataset it is clear that GCSL performs above GPM - but this is not the case with FMNIST. Generally speaking, the classification task on the MNIST dataset can be done with distinct features for each label whereas FMNIST dataset contains very similar or overlapping features for different labels. This could indicate that GCSL is better at learning new features without compromising on previous feature but may not be as capable utilizing previous feature for the current task being learnt. As can be seen in table \ref{FMNIST-comptable}, depending on the architecture either method may lead to a slightly better result. 

\begin{table}[t]
\caption{Accuracy comparison of the GPM vs GCSL methods on the MNIST dataset. The Upper Bound is set by training on all labels at once. GCSL is trained with a V layer configuration at 80\% the size of each hidden layer. The architecture columns represents the size of the hidden layers. The Accuracy on all Labels shows the average ± two standard deviation over the set of experiment iterations.}
\label{MNIST-comptable}
\begin{center}
\begin{tabular}{lll}
\multicolumn{1}{c}{\bf Method}  &\multicolumn{1}{c}{\bf Architecture} &\multicolumn{1}{c}{\bf Accuracy on all Labels}
\\ \hline \\
GPM & [20, 20] & 0.5336 ± 0.0625  \\
GCSL & [20, 20] & 0.7296 ± 0.0538  \\
Upper Bound & [20, 20] & 0.9025 ± 0.0413\\
 \hline \\
GPM & [100, 100] & 0.6505 ± 0.0556  \\
GCSL & [100, 100] & 0.8567 ± 0.0121  \\
Upper Bound & [100, 100] & 0.9375 ± 0.0337\\
\hline \\
GPM & [400, 400] & 0.7292 ± 0.0410  \\
GCSL & [400, 400] & 0.8782 ± 0.0132  \\
Upper Bound & [400, 400] & 0.9521 ± 0.0249\\
\end{tabular}
\end{center}
\end{table}

\begin{table}[t]
\caption{Accuracy comparison of the GPM vs GCSL methods on the FMNIST dataset. 
GCSL is trained with a V layer configuration at 50\% the size of each hidden layer. The architecture columns represents the size of the hidden layers. The Accuracy on all Labels shows the average ± two standard deviation over the set of experiment iterations.}
\label{FMNIST-comptable}
\begin{center}
\begin{tabular}{lll}
\multicolumn{1}{c}{\bf Method}  &\multicolumn{1}{c}{\bf Architecture} &\multicolumn{1}{c}{\bf Accuracy on all Labels}
\\ \hline \\
GPM & [20, 20] & 0.5933 ± 0.0609  \\
GCSL & [20, 20] & 0.5622 ± 0.1190  \\
Upper Bound & [20, 20] & 0.7961 ± 0.0323\\
 \hline \\
GPM & [100, 100] & 0.6924 ± 0.0563  \\
GCSL & [100, 100] & 0.7171 ± 0.0291  \\
Upper Bound & [100, 100] & 0.8127 ± 0.0111\\
\hline \\
GPM & [400, 400] & 0.7471 ± 0.0194  \\
GCSL & [400, 400] & 0.7298 ± 0.0192  \\
Upper Bound & [400, 400] & 0.8179 ± 0.0098\\
\end{tabular}
\end{center}
\end{table}



\section{Summary}
This paper introduce Gradient Correlation Subspace Learning as new alternative in the family of orthogonal gradient based approaches in continual learning. The paper demonstrated the performance of the method on lean networks to highlight its robust ability with remarkably minor collateral (the orthogonal subspaces $V$) from task to task. The high level of customizability around which layers the method is applied to and the easy incorporation into the standard training scheme (it only involves freezing specific weights during training but otherwise trains normally) makes it a realistic contender to be used in a range of applications where there is a serious limit to the amount and size of collateral that can be carried with the network for future learnings. The code for this implementation will be released to GitHub during the publication of this paper.

\section{Discussion}
The GCSL method explored in this paper can be explored to address some pressing topics by being combined with other approaches. GCSL can be applied to fully connected layers on self-supervised networks that were then applied to a specific task and would like to be applied to additional tasks over time. The method could be applied in combination with a replay-based approach in which samples are saved from task to task to help fine tune previous tasks. Furthermore, the saved samples could be incorporated into accumulative correlation matrix stage or potentially incorporated into a contrastive learning scheme to upsample the relatively few samples stored form task to task. 

The experiments in this paper were relatively simple, with a set number of epochs from task to task and a constant subspace layer configuration from task to task. Further exploration is possible by incorporating early stopping or dynamically setting the subspace layer configuration from task to task. Throughout the experiments, it was visible that the more tasks that have been learned the slower the next task learns. These considerations can shape the subspace layer configuration and the eigenvalues of the subspace can also be taken into a deeper consideration. This paper did not explore utilizing the GCSL method on CNN architecture and that remains an area for future exploration.

The comparison experiments showed that this method can perform on-par or sometimes above other state-of-the-art methods but a deeper evaluation is possible. This paper served as an introduction to the topic but there remains a lot to be explored to fully understand the implications and the possibilities of this method.

\bibliography{CoLLAs2022}
\bibliographystyle{CoLLAs2022}


\end{document}